\documentclass[10pt,twocolumn,letterpaper]{article}

\usepackage[pagenumbers]{cvpr}
\usepackage[dvipsnames]{xcolor}
\usepackage{multicol}
\usepackage{multirow}
\usepackage{times}
\usepackage{epsfig}
\usepackage{graphicx}
\usepackage{amsmath}
\usepackage{amssymb}
\usepackage{booktabs}
\usepackage{multicol}
\usepackage{multirow}
\usepackage{rotating}
\usepackage{multirow}
\usepackage{array}
\usepackage{booktabs}

\usepackage{lipsum}
\usepackage{colortbl}
\usepackage{footmisc}
\usepackage{setspace}

\usepackage{pifont}%
\newcommand{\xmark}{\ding{55}}%
\newcommand{\cmark}{\ding{51}}%
\usepackage[numbers]{natbib}

\definecolor{cvprblue}{rgb}{0.21,0.49,0.74}
\usepackage[pagebackref,breaklinks,colorlinks,citecolor=cvprblue]{hyperref}

\def\paperID{5} %
\def\confName{CVPR}
\def\confYear{2024}

\title{CONDA: Continual Unsupervised Domain Adaptation Learning in \\ 
Visual Perception for Self-Driving Cars}

\author{
Thanh-Dat Truong$^1$, Pierce Helton$^1$, Ahmed Moustafa$^1$, Jackson David Cothren$^2$, Khoa Luu$^1$ \\
$^1$CVIU Lab, University of Arkansas, Fayetteville, AR, USA \\
$^2$Department of Geosciences, University of Arkansas, Fayetteville, AR, USA \\
{\tt\small \{tt032, pchelton, armoustam, jcothre, khoaluu\}@uark.edu}
}

\begin{document}
\maketitle

\begin{abstract}
Although unsupervised domain adaptation methods have achieved remarkable performance in semantic scene segmentation, these approaches remain impractical in real-world use cases. In practice, the segmentation models may encounter new data that have not been seen yet. Also, the previous data training of segmentation models may be inaccessible due to privacy problems. Therefore, to address these problems, in this work, we propose a \textbf{Co}ntinual Unsupervised \textbf{D}omain \textbf{A}daptation (CONDA) approach that allows the model to continuously learn and adapt with respect to the presence of the new data. Moreover, our proposed approach is designed without the requirement of accessing previous training data. To avoid the catastrophic forgetting problem and maintain the performance of the segmentation models, we present a novel Bijective Maximum Likelihood loss to impose the constraint of predicted segmentation distribution shifts. The experimental results on the benchmark of continual unsupervised domain adaptation have shown the advanced performance of the proposed CONDA method.
\end{abstract}

\section{Introduction}

Semantic scene segmentation has become one of the most popular research topics in computer vision recently. Its goal is to break down an input image and densely assign each pixel to its corresponding predefined class. There have been new approaches recently based upon deep learning technology that have remarkable results in semantic scene segmentation. \cite{chen2018deeplab}. 
Typically, segmentation is trained on labelled scene data, but annotating the images for semantic segmentation is a time-consuming and expensive process. This is because each and every pixel in the input image must be labelled. A method to reduce the cost of labelling images is to use a simulation to create a large-scale synthetic dataset \cite{Richter_2016_ECCV, Ros_2016_CVPR}. Although this strategy saves time, it has a serious trade off when deploying supervised models trained on these synthetic datasets into the real-world data. Particularly, these supervised models trained on the synthetic datasets often perform worse on real images due to a pixel appearance gap between the synthetic and real images and thus are not well-suited for real-image deployment.

\begin{figure}
    \centering
    \includegraphics[width=0.48\textwidth]{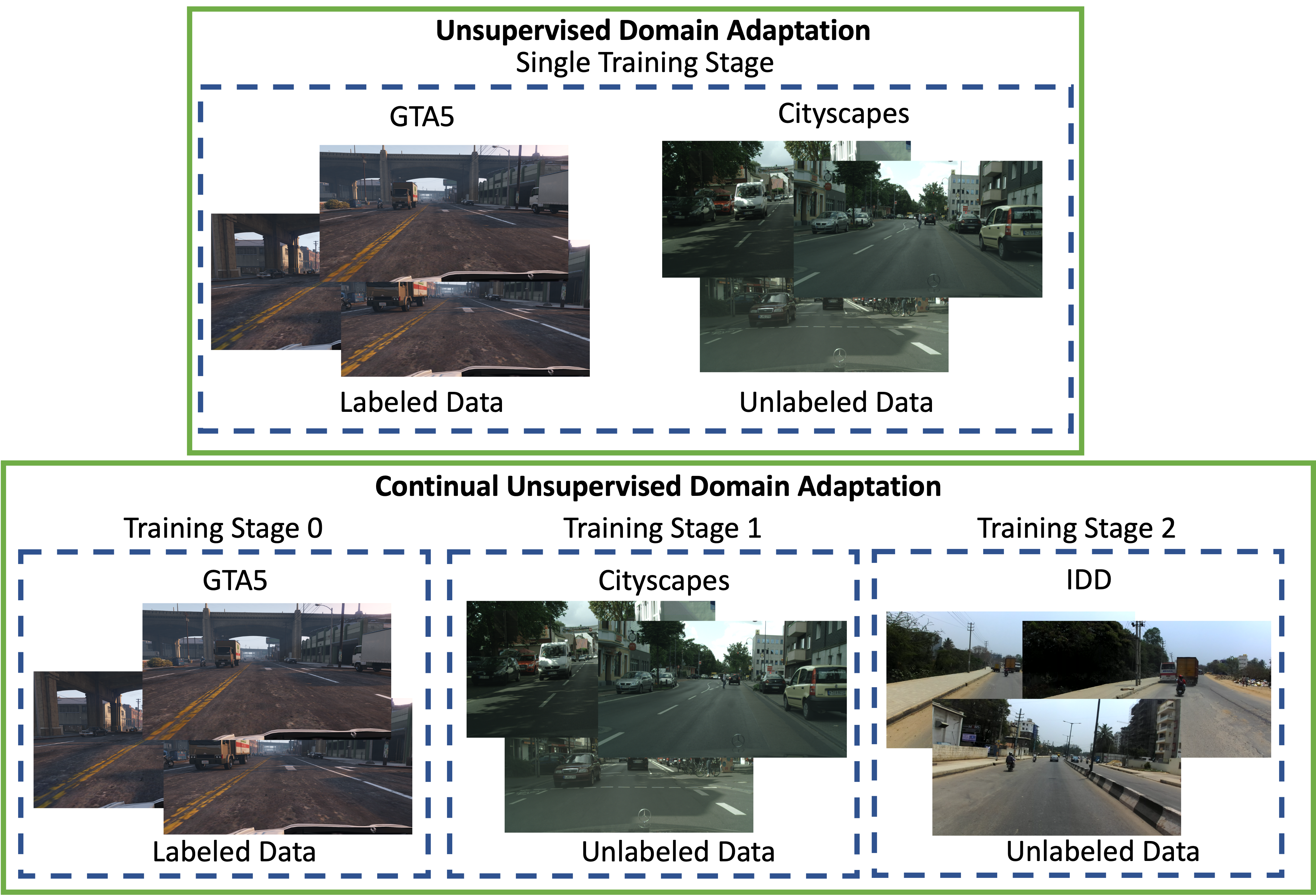}
    \caption{Unsupervised domain Adaptation trains the model on both labeled source data and unlabeled target data simultaneously by a single training stage. Meanwhile, Continual Unsupervised Domain Adaptation first trains on the labeled source data. Then, it continuously adapts the model to the new target domains and at each training stage, the model only accesses data at that stage.}
    \label{fig:uda_cuda}
\end{figure}

\begin{table*}[t]
\footnotesize
\centering
\caption{ \textbf{Comparisons in the properties between our proposed approach and other prior methods}. Convolutional Neural Network (CNN), Generative Adversarial Net (GAN), Bijective Network (BiN), Entropy Minimization (EntMin), Segmentation Map (Seg), Depth Map (Depth); $\ell_{IW}$: Image-wise Weighting Loss, $\ell_{CE}$: Cross-entropy Loss, $\ell_{Focal}$: Focal Loss $\ell_{adv}$: Adversarial Loss, $\ell_{Huber}$: Huber Loss. $\ell_{square}$: Maximum Squares Loss,  $\ell_{density}$: Negative Log-Likelihood Loss, $\ell_{bml}$:  Bijective Maximum Likelihood Loss.}
\begin{tabular}{|c|c|c|c|c|c|c|c|}
\hline
\textbf{Methods} & \begin{tabular}{@{}c@{}} \textbf{Source-Free}\\\textbf{Data}  \end{tabular}  & \begin{tabular}{@{}c@{}} \textbf{Forgetting} \\ {\textbf{Aware}}\end{tabular}           & \begin{tabular}{@{}c@{}}\textbf{Structural}\\\textbf{Learning} \end{tabular} & \begin{tabular}{@{}c@{}}\textbf{Source} \\ \textbf{Label Require}\end{tabular} & \begin{tabular}{@{}c@{}} {\textbf{Target Domain}} \\ {\textbf{Transfer}} \end{tabular} & \textbf{Architecture}& \textbf{Designed Loss}  \\ 
\hline
AdaptPatch \cite{tsai2019domain} & \xmark& \xmark& Weak (Binary label) & Seg  & \cmark & CNN+GAN & $\ell_{adv}$ \\
CBST \cite{zou2018unsupervised} &\xmark & \xmark & $-$ & Seg & \cmark & CNN & $\ell_{CE}$\\
ADVENT \cite{vu2019advent} &\xmark & \xmark & Weak (Binary label) & Seg  & \cmark & CNN+GAN & $\operatorname{EntMin} + \ell_{adv}$ \\
IntraDA \cite{pan2020unsupervised} & \xmark& \xmark & Weak (Binary label) & Seg & \cmark & CNN+GAN & $\operatorname{EntMin} + \ell_{adv}$ \\
BiMaL \cite{truong2021bimal}        & \xmark  & \xmark         & Maximum Likelihood   & Seg    & \cmark        & \begin{tabular}{@{}c@{}} CNN + BiN \end{tabular} & $\ell_{density}$ \\
SPIGAN \cite{lee2018spigan}  &\xmark & \xmark & Weak (Binary label) & Seg + Depth & \cmark & CNN+GAN & $\ell_{adv}$ \\
DADA \cite{vu2019dada} &\xmark & \xmark & Depth-aware Label & Seg + Depth & \cmark & CNN+GAN & $\ell_{adv}+\ell_{Huber}$ \\
MaxSquare \cite{chen2019domain} & \xmark & \xmark & \begin{tabular}{@{}c@{}} Weak (Binary label) \end{tabular} & Seg & \cmark & CNN + GAN & $\ell_{square} + \ell_{IW}$\\
SAC \cite{Araslanov:2021:DASAC} & \xmark & \xmark & $-$ & Seg & \cmark & CNN & $\ell_{CE} + \ell_{Focal}$\\
\hline \hline
\textbf{Ours} & \cmark & \cmark   & \begin{tabular}{@{}c@{}} \textbf{Bijective} \\ \textbf{Maximum Likelihood} \end{tabular}   &   $-$  & \cmark  & \begin{tabular}{@{}c@{}} \textbf{CNN + BiN} \end{tabular} &  $\ell_{bml}$ \\ \hline
\end{tabular}
\label{tab:summary}
\end{table*}

To address the problem aforementioned, unsupervised domain adaptation has emerged as a feasible solution. In particular,
unsupervised domain adaptation (UDA) aims to learn a model on the large-scale annotated source datasets (referred as the source domain) and adapt to unlabeled target datasets (referred as the target domain) to guarantee its performance on the new domain. 
Common unsupervised domain adaptation methods try to minimize the distribution discrepancy in the deep representations between the source and target domains. The minimization process is done simultaneously with supervised learning on the source domain \cite{truong2021bimal, tsai2018learning, tsai2019domain, vu2019advent, vu2019dada}.
The discrepancy minimization can computed at a single level or multiple levels of deep representation using maximum mean discrepancy \cite{ganin2015unsupervised, long2015learning}, or adversarial loss within a generative adversarial network framework \cite{chen2018road, hoffman18a, hong2018CVPR, tsai2018learning, tzeng2017adversarial}, or contrastive learning \cite{kang2019contrastive, Yue_2021_CVPR}. In addition, other domain adaptation approaches utilize the cross entropy loss with pseudo labels with well designed learning strategies \cite{Araslanov:2021:DASAC, zhang2021prototypical, zou2018unsupervised}. 

These approaches have shown their potential performance in semantic segmentation with domain adaptation; however, their practical applications are limited. For example, in autonomous driving, vehicles may encounter various urban or highway scenarios with a diversity of environments, e.g., weather, lighting, or geographical conditions, and each of these conditions can be considered as a new target domain. 
Meanwhile, in the domain adaptation approach, the model is trained on a particular given target domain not able to update itself with the presence of new data. Moreover, the training data, in some cases, is protected during the adaptation phase to preserve privacy.
In this work, we address this problem by introducing a continual learning framework for semantic scene segmentation under the domain adaptation setting. The goal of our proposed framework is to sequentially, continuously learn and adapt the model to the new incoming unlabeled target domain while maintaining the performance of the model on previous target domains and alleviating the catastrophic forgetting problem. In contrast to prior domain adaptation approaches, our continual unsupervised domain adaptation does not access the previous training data during the adaptation phase. In other words, only the unlabeled target data is given during the adaptation phase. Figure \ref{fig:uda_cuda} illustrates the difference between unsupervised domain adaptation and continual unsupervised domain adaptation.
Table \ref{tab:summary} summaries the difference between our proposed approach and prior standard domain adaptation approaches. 

\noindent
\textbf{Contributions of This Work}
In summary, this work presents a novel Continual Unsupervised Domain Adaptation (CONDA) approach to semantic scene segmentation.
The contributions in this work can be summarized as follows. Firstly, the problem of continual domain adaptation is formulated in semantic scene segmentation by regularizing the distribution shift of predictions between source and target domains to avoid the catastrophic forgetting problem. As opposed to prior adaptation methods, only data from the target domains are given during the adaptation phase in our continual learning framework. Secondly, given the formulated problem, the distribution shift regularizer is further derived into the Bijective Maximum Likelihood loss that can be used to measure the distribution shift without the demand of the source training data. Then, the bijective maximum likelihood loss is formed using the bijective network so that the loss is able to capture global structural information and represents the distributions of the segmentation.
Finally, though our experiments on the benchmark of GTA $\to$ Cityscapes $\to$ IDD $\to$ Mapillary, our proposed approach outperforms the prior methods and achieves state-of-the-art results.

\section{Related Work}

\subsection{Semantic Segmentation via Deep Learning} 

Fully Convolutional Networks (FCNs) are the preferred approach for the task of semantic segmenation due to their capacity for learning and high accuracy. Combining FCNs with an encoder-decoder structure produces additional refinements in accuracy. 
FCNs \cite{long2015fully, chen2018deeplab} were first applied to the task of segmentation with multiple convolutional layers followed by spatial pooling. Later approaches \cite{lin2017refinenet, pohlen2017full} combined upsampled, high-level feature maps with low-level feature maps prior to decoding, collecting more information and still producing precise instance borders.
Additional works \cite{chen2018deeplab, DBLP:journals/corr/YuK15} improve the performance of models while preserving the field of view by using  dilated convolutions. Spatial pyramid pooling has seen success in other recent developments \cite{chen2017rethinking, chen2018encoder}. This method allows the model to acquire contextual information at multiple levels, generating more global information at higher network layers.
In a novel FCN architecture, Deeplabv3+ \cite{chen2017rethinking} used the encoder-decoder structure in unison with spatial pyramid pooling to produce a faster, stronger network.
The latest works use Transformer-based backbones \cite{xie2021segformer, daformer, transda} to create more proficient and advanced semantic segmentation networks.

\subsection{Unsupervised Domain Adaptation via Deep Learning}

The common domain adaptation approaches are domain discrepancy minimization \cite{ganin2015unsupervised, long2015learning, tzeng2017adversarial}, adversarial learning \cite{chen2018road,chen2017no, hoffman18a, hoffman2016fcns, hong2018CVPR, tsai2018learning}, entropy minimization \cite{murez2018CVPR, pan2020unsupervised, vu2019advent, zhu2017unpaired}, self-training \cite{zou2018unsupervised, Truong:CVPR:2023FREDOM, hoyer2022hrda}.
Within this work's scope, UDA is focused on semantic segmentation, and adversarial training is the most commonly employed approach to UDA for semantic segmentation. The adversarial training paradigm is much like generative adversarial networks (GANs) in that they both aim to train a predictive discriminator on the domain of inputs while the segmentation network tries to fool the discriminator. 
Both the training of the adversarial step and the supervised segmentation task occur simultaneously on the source and target domains.
The first instance of a GAN-based UDA approach to segmentation was introduced by Hoffman \etal \cite{hoffman2016fcns}. Later, global and class-wise adaptation learned from the application of adversarial learning on pseudo labels was presented by Chen \etal \cite{chen2017no}. After considering the spatial distribution difference, \cite{chen2018road} proposed a spatial-aware adaptation method to align two domains along with a target-guided distillation loss. A conditional generator that transforms feature maps of the source domain to better match the target domain was learned by Hong \etal \cite{hong2018CVPR}.
Tasi \etal \cite{tsai2018learning} learned a consistency of scene layout and local context between target and source domains by using adversarial learning. There exist prior methods that utilize GANs to use source images %
to synthesize target images \cite{zhu2017unpaired, murez2018CVPR}. Hoffman \etal \cite{hoffman18a} presented Cycle-Consistent Adversarial Domain Adaptation which aligns at both the pixel and feature level representations. Zhu \etal \cite{zhu2018ECCV} developed the penalization of easy and hard source examples by implementing a conservative loss in the adversarial framework. We \etal \cite{wu2018dcan} proposed a DCAN framework that uses channel-wise feature alignment in the segmentation networks.
Sakaridis \etal \cite{SDHV18} proposed a UDA framework on scene understanding that can gradually adapt a segmentation model for increasingly foggy images, i.e. no fog to high fog.
Recently, self-supervised approaches \cite{zhang2021prototypical, Araslanov:2021:DASAC, daformer} have shown their state-of-the-art performance in domain adaptation tasks.
Araslanov \etal \cite{Araslanov:2021:DASAC} developed a simple self-supervised framework trained on pseudo labels without the demand of extra training rounds.
Zhang \etal \cite{zhang2021prototypical} introduced a self-training approach that is able to denoise pseudo labels and learns structural information by enforcing the consistency between augmentations. 
Hoyer \etal \cite{daformer} improved the performance of segmentation models by utilizing pseudo labels and introducing a powerful transformer-based backbone. 
Later, Hoyer \etal further improved their approach by using multi-resolution cropped images \cite{hoyer2022hrda} and masked image consistency learning strategy \cite{hoyer2023mic} to enhance contextual learning.
Fashes et al. \cite{fahes2023poda} presented a new prompt-based approach to zero-shot unsupervised domain adaptation.

\subsection{Continual Learning} 

These methods aim to continuously update the model with respect to an incoming streaming data.
The simple approach to this task is to re-train or fine-tune the model with the new data.
However, under the assumption that the original training data is inaccessible due to the privacy problem, fine-tuning the model on the new updated data could lead to the problem of catastrophic forgetting \cite{robins1995catastrophicforgetting,thrun1998lifelonglearning,french1999catastrophicforgetting, kirkpatrick2017ewc, lopezpaz2017gem, douillard2020podnet, douillard2021plop, ozdemir2018learn, ozdemir2019extending, michieli2019incremental} which is the most challenging problem in continual learning. Catastrophic forgetting refers the existence of performance drops on the original data when the model is updated.
This challenge can be alleviated by constraining the updated model to a similar locale of the previous model. This constraint can be done by imposing the regularization on gradients or weights \cite{kirkpatrick2017ewc, lopezpaz2017gem}, or measuring the probabilities \cite{li2018lwf}, or distilling the intermediary features \cite{douillard2020podnet}.
Recently, several works \cite{michieli2019ilt,cermelli2020modelingthebackground} developed the continual learning framework for semantic segmentation. However, these approaches are not applicable for the unsupervised domain adaptation setting as these requires the ground truth labels of data during the learning process.
\cite{volpi2021continual} presented a continual learning framework where the model is sequentially trained on multiple labeled data domains. 
\cite{rostami2021lifelong} proposed a continual learning framework under the unsupervised domain adaptation setting. However, this approach requires storing all data for the purpose of rehearsal.
\cite{saporta2022muhdi} introduced a multi-dead knowledge distillation framework for the continual unsupervised domain but this approach requires access to the source data during the training phase.
The recent approaches presented a clustering approach to continual learning \cite{joseph2021towards, truong2023fairness}
In our work, we focus on the continual learning framework under the unsupervised adaptation setting where the source data is not used in the adaptation process.

\section{The Proposed CONDA Framework}

\begin{figure*}
    \centering
    \includegraphics[width=1.0\textwidth]{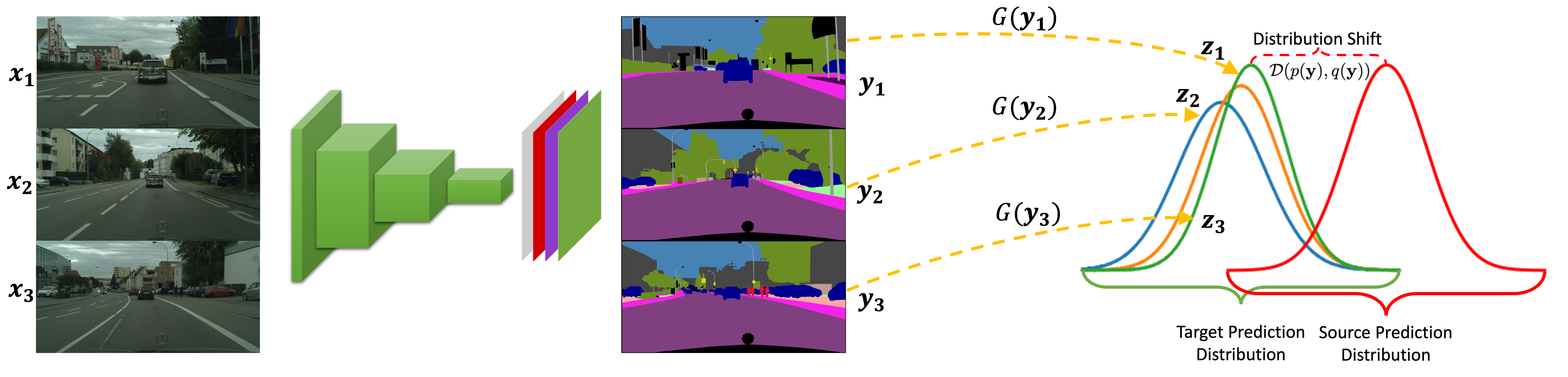}
    \caption{\textbf{The proposed CONDA Framework.} The input images $\mathbf{x}$ are first forward to the segmentation network $F$. Then, the segmentation maps $\mathbf{y}$ are projected into the deep latent space by the bijective network $G$ to measure the distribution shift of target predictions compared to the original distribution of source predictions.}
    \label{fig:framework}
\end{figure*}

In this section, we present the proposed domain adaptation approach in the context of continual learning.
Let $\mathbf{x}_s \in \mathcal{X}_s$ be the RGB input image of the source domain, $\mathbf{\hat{y}}_s \in \mathcal{Y}_s$ be the corresponding segmentation ground truth of $\mathbf{x}_s$, and
$F$ be the deep network that maps the input image $\mathbf{x}$ into the segmentation $\mathbf{y}$, i.e $\mathbf{y} = F(\mathbf{x}, \theta)$, $\theta$ is the deep network parameters. We assume that the network $F$ has been first well learned on the source domain by supervised learning defined as follows,
\begin{equation}
    \theta^* = \arg\min_{\theta}\mathbb{E}_{\mathbf{x}_s, \mathbf{y}_s \in \mathcal{X}_s \times \mathcal{Y}_s}\mathcal{L}_{s}(F(\mathbf{x}_s, \theta), \mathbf{\hat{y}}_s) 
\end{equation}
where $\mathcal{L}_s$ is the supervised loss, i.e., the cross-entropy loss, between a predicted segmentation and a ground-truth segmentation. 
Then, given a well-learned $F$, we will continuously learn the model on the incoming target data. 
\textbf{In our setting, we assume that the given source dataset is not used to train during the adaptation phase.}
Let $\mathbf{x} \in \mathcal{X}_t$ be the RGB input image of the incoming target domain $\mathcal{X}_t$. 
We assume there could be multiple incoming target datasets that the model needs to continually adapt to. 
Let $q(\mathbf{y})$ represent the distribution of predicted segmentation of the previous model learned on the source domain, $p(\mathbf{y})$ be the distribution of predicted segmentations of the current updated model.

In general, domain adaptation  in the context of continuous learning of the semantic segmentation task can be formed as follows,
\begin{equation} \label{eqn:DA_KL_Constraint}
\begin{split}
    \theta^* = \arg\min_{\theta}\left[\mathbb{E}_{\mathbf{x} \sim \mathcal{X}_t}\mathcal{L}_t(\mathbf{y}) + \lambda \mathcal{D}(p(\mathbf{y}), q(\mathbf{y}))\right]
\end{split}
\end{equation}
where $\lambda$ is the hyper-parameter imposing the impact of the distribution shift, 
$\mathcal{L}_t$ is the unsupervised loss used in the domain adaptation setting (e.g. entropy loss \cite{vu2019advent}, cross-entropy loss with pseudo labels \cite{Araslanov:2021:DASAC}, adversarial loss \cite{vu2019dada}), and $\mathcal{D}(p(\mathbf{y}),  q(\mathbf{y}))$ is the distance between distributions of source predictions and target predictions, respectively. 
As shown in Eqn. \eqref{eqn:DA_KL_Constraint}, the constraint $\mathcal{D}(p(\mathbf{y}), q(\mathbf{y}))$ ensures that the unsupervised adaptation procedure does not lead to catastrophic forgetting. 
In particular, while the model continuously updates with respect to the incoming target data, it must not forget the knowledge that has been learned in previous training using the source data. This is ensured by imposing the distribution shift between source prediction and target prediction distributions. 
Figure \ref{fig:framework} illustrates our proposed framework.

\subsection{Distribution Shift Modeling by Bijective Network}

There are several options for $\mathcal{D}$ to measure the divergence between the two distributions $p(\mathbf{y})$ and $q(\mathbf{y})$. In our paper, we define distance $\mathcal{D}$ as the Kullback–Leibler (KL) divergence which is a common statistical distance shown in Eqn. \eqref{eqn:KL_diver}.

\begin{equation} \label{eqn:KL_diver}
\begin{split}
\small
    \mathcal{D}\left(p(\mathbf{y}),q(\mathbf{y})\right) &= \int \log\left(\frac{p(\mathbf{y})}{q(\mathbf{y})}\right)p(\mathbf{y})d\mathbf{y} \\
\end{split}
\end{equation}
Eqn. \eqref{eqn:KL_diver} measures the divergence between distributions $p(\mathbf{y})$ and $q(\mathbf{y})$. However, Eqn. \eqref{eqn:KL_diver} requires access to the source dataset during the adaptation phase which conflicts with the setting in our problem.
Thus, Eqn. \eqref{eqn:KL_diver} is re-formed as follows,
\begin{equation} \label{eqn:KL_diver_upper}
\begin{split}
    & \int \log\left(\frac{p(\mathbf{y}_t)}{q(\mathbf{y}_t)}\right)p(\mathbf{y})d\mathbf{y} \\
    &= {\mathbb{E}}_{\mathbf{y} \sim p(\mathbf{y})} \log(p(\mathbf{y}))  - {\mathbb{E}}_{\mathbf{y} \sim p(\mathbf{y})} \log(q(\mathbf{y})) \\
    &\leq - {\mathbb{E}}_{\mathbf{y} \sim p(\mathbf{y})} \log(q(\mathbf{y}))
\end{split}
\end{equation}
With any form of the distribution $p$, the above inequality always holds as $p(\mathbf{y}) \in [0,1]$ and $\log(p(\mathbf{y})) \leq 0$. Then, instead of directly computing the KL divergence between $p(\mathbf{y})$ and $q(\mathbf{y})$, we define $\mathcal{D}_{MaL}$ as the upper bound of Eqn. \eqref{eqn:KL_diver},
\begin{equation}\label{eqn:loss_for_D}
    \mathcal{D}_{MaL}\left(p(\mathbf{y}),q(\mathbf{y})\right) = - {\mathbb{E}}_{\mathbf{y} \sim p(\mathbf{y})} \log(q(\mathbf{y}))
\end{equation}

As Eqn. \eqref{eqn:loss_for_D} is an upper bound of Eqn. \eqref{eqn:KL_diver} due to Eqn. \eqref{eqn:KL_diver_upper}. Then, minimizing Eqn. \eqref{eqn:loss_for_D} also imposes the distance between two distributions  $p(\mathbf{y})$ and $q(\mathbf{y})$.

As $q(\mathbf{y})$ represents the distribution of predicted segmentation of the model learned on the source dataset, 
we propose to model the distribution $q(\mathbf{y})$ by the bijective network. Let $G: \mathbf{Y}_S \to \mathcal{Z}$ be the bijective network that maps a segmentation to the latent space, i.e. $\mathbf{z} = G(\mathbf{y}, \theta_G)$, where $\mathbf{z} \sim \pi(\mathbf{z})$ is the latent variable, $\theta_G$ is the set parameters of $G$, and $\pi$ is the prior distribution.

Then, by the change of variable formula, the distribution $p(\mathbf{y})$ can be formed as follows,
\begin{equation} \label{eqn:bijective_mapping}
\small
    \log p(\mathbf{y}) = \log \pi(\mathbf{z}) + \log\left|\frac{\partial G(\mathbf{y}, \theta_G)}{\partial \mathbf{y}}\right|,
\end{equation}

where $\left|\frac{\partial G(\mathbf{y}, \theta_G)}{\partial \mathbf{y}}\right|$ is the Jacobian determinant of $G(\mathbf{y}, \theta_G)$ with respect to $\mathbf{y}$. 
Then, the network $G$ can be learned by minimizing the negative log-likelihood on the source segmentation maps as follows,

\begin{equation} \label{eqn:BijectiveLearning}
\footnotesize
\begin{split}
    \theta_G^{*} =& \arg\min_{\theta_G} %
    \mathbb{E}_{\mathbf{y} \in \mathcal{Y}_s} \Big[-\log(q(\mathbf{y}))\Big] \\
    =& \arg\min_{\theta_G} \mathbb{E}_{\mathbf{y} \in \mathcal{Y}_s} \left[ \log \pi(\mathbf{z}) + \log\left|\frac{\partial G(\mathbf{y}, \theta_G)}{\partial \mathbf{y}}\right| \right].
    \raisetag{40pt}
\end{split}
\end{equation}
Generally, there are several choices for the prior distribution $\pi$. However, the ideal prior distribution should meet two conditions: (i) simplicity in the density estimation, and (ii) easy in sampling.
Therefore, taking these conditions into account, we adopt the the Normal distribution as the prior distribution $\pi$.
It should be noted that the prior distribution is not limited to the Normal distribution; any form of distribution that satisfies the two conditions can also be adopted.

To enhance the modeling capability of the bijective networks $G$, we design $G$ as a multi-scale architecture network in which each scale is designed as an invertible network.
Several deep neural structures \cite{dinh2015nice, dinh2017density,Duong_2017_ICCV, duong2020vec2face, glow, duong2019learning, truong2021fastflow} can be adopted to construct the invertible network at each scale. We assume that the bijective network $G$ and the segmentation network $F$ are given during the adaptation phase.

\section{Experimental Results}

In this section, we first review the datasets used in our experiments. Then, we describe our implementation in detail and the evaluation metric used to measure the performance of the segmentation models. Next, we present our experimental results on the standard benchmark of continual unsupervised domain adaptation. Finally, we compare our qualitative results with the baselines. 

\begin{table*}[ht!]
    \centering
            \begin{tabular}{c| c | c| c c c c c c c|c|c<{\kern-\tabcolsep}}
                \toprule
                 & Method & Target  & \rotatebox{90}{flat} & \rotatebox{90}{constr.} & \rotatebox{90}{object} & \rotatebox{90}{nature} & \rotatebox{90}{sky} & \rotatebox{90}{human} & \rotatebox{90}{vehicle\,} & \rotatebox{90}{mIoU}&\rotatebox{90}{\shortstack[c]{mIoU\\Avg.}}\\
                \midrule
                \midrule
                \multirow{9}{*}{\rotatebox[origin=c]{90}{\shortstack[c]{Multi-Target\\Oracle Setting}}}&\multirow{3}{*}{\shortstack[c]{Multi-Target\\Baseline \cite{vu2019advent}}} & Cityscapes &  93.6 &	80.6 &	26.4 &	78.1 &	81.5 &	51.9 &	76.4 &	69.8 & \multirow{3}{*}{67.8}  \\
                & & IDD &  92.0 &	54.6 &	15.7 &	77.2 &	90.5 &	50.8 &	78.6 &	65.6 & \\
                & & Mapillary &  89.2 &	72.4 &	32.4 &	73.0 &	92.7 &	41.6 &	74.9 &	68.0 & \\
                \cmidrule{2-12}
                &\multirow{3}{*}{Multi-Dis. \cite{saporta2021mtaf}}  & Cityscapes &  94.6 &	80.0 &	20.6 &	79.3 &	84.1 &	44.6 &	78.2 &	68.8 & \multirow{3}{*}{68.2}\\
                & & IDD &  91.6 &	54.2 &	13.1 &	78.4 &	93.1 &	49.6 &	80.3 &	65.8 &  \\
                & & Mapillary &  89.0 &	72.5 &	29.3 &	75.5 &	94.7 &	50.3 &	78.9 &	70.0 & \\
                \cmidrule{2-12}
                &\multirow{3}{*}{MTKT \cite{saporta2021mtaf}} &Cityscapes &  94.6 &	80.7 &	23.8 &	79.0 &	84.5 &	51.0 &	79.2 &	70.4 & \multirow{3}{*}{69.1} \\ 
                & & IDD &  91.7 &	55.6 &	14.5 &	78.0 &	92.6 &	49.8 &	79.4 &	65.9 &   \\
                & & Mapillary &  90.5 &	73.7 &	32.5 &	75.5 &	94.3 &	51.2 &	80.2 &	71.1 & \\
                
                \midrule
                \midrule
                \multirow{6}{*}{\rotatebox[origin=c]{90}{\shortstack[c]{Continual\\Setting}}}&\multirow{3}{*}{\shortstack[l]{Continual\\Baseline \cite{vu2019advent}}}  & Cityscapes & \textbf{92.9} & \textbf{79.0} & 18.7 & \textbf{76.9} & 84.1 & 47.3 & 72.9 & 67.4  & \multirow{3}{*}{67.0}  \\
                & & IDD & \textbf{91.8} & 51.1 & 11.6 &\textbf{79.0} &91.6 &\textbf{47.5} &\textbf{72.5} &63.6 & \\
                & & Mapillary &  90.3 & 71.7 & 30.1 & 76.1 & \textbf{93.9} & 50.2 & 77.3 & 70.0 & \\

                \cmidrule{2-12}
                &  & Cityscapes &  89.4	& {71.1}	& \textbf{32.8}	& 75.8	& \textbf{92.8}	& \textbf{48.1}	& \textbf{75.6}	& \textbf{69.4} &   \\
                & & IDD &  88.0	& \textbf{69.4}	& \textbf{30.6}	& 73.0	& \textbf{92.8}	& 47.1	& 62.6	& \textbf{66.2} & \textbf{68.9} \\
                &\multirow{-3}{*}{Ours}& Mapillary &  \textbf{90.4}	& \textbf{73.3}	& \textbf{33.1}	& \textbf{76.4}	& 93.8	& \textbf{51.2} & \textbf{80.1} & \textbf{71.2} & \multirow{-3}{*}{\textbf{-}}\\
                
                \bottomrule
            \end{tabular}
    \vspace{0.2cm}
    \caption{\textbf{Segmentation performance (\%) on GTA5 $\to$ Cityscapes $\to$ IDD $\to$ Mapillary.}}
    \label{tab:ctkt_results}
    \end{table*}

\subsection{Datasets and Network Architectures}

\noindent
\textbf{GTA5} is a collection of synthetic images and semantic labels at a resolution of $1914 \times 1052$ pixels. The $24,966$ images were collected from the GTA5 game engine using the communication between the game engine and the graphics hardware. In total, 33 class categories are included in the dataset, but only 7 categories are compatible with the Cityscapes \cite{cordts2016cityscapes}, IDD \cite{Varma2019IDDAD}, and Mapillary  \cite{MVD2017}  datasets used in our experiments.

\noindent
\textbf{Cityscapes} includes $3,975$ images captured from 50 different cities. Each image has a high-quality, semantic, dense pixel annotation of 30 object classes from the urban settings.
Cityscapes was created to give researchers more annotated, high quality, and real world data.

\noindent
\textbf{Vistas (Mapillary Vistas Dataset)} \cite{MVD2017} is a dataset of various street-level images with pixel‑accurate and instance‑specific human annotations from areas throughout the world. The data includes $25,000$ high-quality images and $124$ object categories. 

\noindent
\textbf{IDD} \cite{Varma2019IDDAD} contains images from several cities in India. The dataset has $10,000$ semantically labeled images. Of the 34 object classes included in the dataset, only 7 classes that are commmon to SYNTHIA, Cityscapes, and Vistas are used in our experiments.

\noindent
\textbf{Implementation} The DeepLab-V2 \cite{chen2018deeplab} architecture with a Resnet-101 \cite{He2015} backbone is used in all experiments. In our experiments, the image size is set to the resolution of $640 \times 320$. We use Exponential Moving Averages (EMA) to update the model. 
In our work, the unsupervised loss $\mathcal{L}_t$ is defined as the cross-entropy loss with the pseudo labels computed by the EMA model. The value of $\lambda$ is set to $0.005$.

\noindent
We adopt the structure of \cite{glow, truong2021bimal, duong2020vec2face} for our generator $G$. 
We used PyTorch \cite{paszke2019pytorch} to implement the framework. 4 NVIDIA Quadpro P800 GPUs with 48GB of VRAM each were utilized during development. 
The model is optimized by the Stochastic Gradient Descent optimizer where the learning rate is set to $2.5 \times 10^{-4}$.

\noindent
\textbf{Evaluation Metric} The performance of semantic segmentation models are often measured by the mean Intersection over Union (mIoU) metric over all classes, expressed in percentage. 
Follow the evaluation protocol of prior domain adaptation methods \cite{vu2019advent, vu2019dada, truong2021bimal, Araslanov:2021:DASAC}, we also adopt the mIoU metric in our experiments to measure the performance of different methods.

\subsection{Quantitative Results}

In our experiments, we compare our methods with the baseline (AdvEnt) \cite{vu2019advent}, Multi-Dis. \cite{saporta2021mtaf}, and MTKT \cite{saporta2021mtaf} on the benchmark of GTA5 $\to$ Cityscapes $\to$ IDD\ $\to$ Mapillary. 
There are two setting in our experiments, i.e., (1) The Multi-target Oracle setting means the model is simultaneous training on all the target domains, and  (2) The Continual setting means that the model is trained target domains sequentially. It should be noted that the baseline (AdvEnt) use the source data during the adaptation phase. 

\textbf{GTA $\to$ Cityscapes:} 
Table \ref{tab:ctkt_results} displays our method's SOTA performance when compared with prior methods on 7 shared classes from the Cityscapes validation set. In particular, our approach achieved a higher IoU than the previous baseline by $+1.9\%$. Comparing the per class results with the baseline, 3 of the 7 classes saw notable improvements, namely \textit{``object''} ($+14.1\%$), \textit{``sky''} ($+8.7\%$), and \textit{``vehicle''} ($+2.9\%$). The remaining classes did not differ from the baseline by a large degree.

\textbf{\textbf{GTA $\to$ Cityscapes $\to$ IDD:}}
In the IDD row of Table \ref{tab:ctkt_results}, the improvements for the 7 shared classes can be seen when adapting from GTA5 $\to$ Cityscapes $\to$ IDD.
Again, our approach improves the baseline IoU results, with a per class improvement in IoU of $18.3\%$, $19.0\%$, and $1.2\%$ for the classes of \textit{``constr''}, \textit{``object''}, and \textit{``sky''}, respectively. 
Furthermore, the overall mIoU accuracy improved by $2.6\%$ from the previous baseline.

\textbf{GTA $\to$ Cityscapes $\to$ IDD $\to$ Mapillary:} 
The final experiment tested the adaptation of the model to the Mapillary dataset. Table \ref{tab:ctkt_results} displays the results of this experiment.
Our approach improved the previous baseline's mIoU of all classes by an average of $1.2\%$. The improvement by class was $0.1\%$ for \textit{``flat''}, $1.6\%$ for \textit{``constr''}, $3.0\%$ for \textit{``object''}, $0.3\%$ for \textit{``nature''}, $1.0\%$ for \textit{``sky''}, and $2.8\%$ for \textit{``vehicle''}. Across all experiments, our new approach improved the mIoU average by $1.9\%$, setting a new SOTA result for the task of continuous learning. It should be noted that without accessing the source training data during the adaptation phase, overall, our approach still gains better results compared to the baseline method.

\subsection{Qualitative Results}

\begin{figure}[t]
    \centering
    \includegraphics[width=0.48\textwidth]{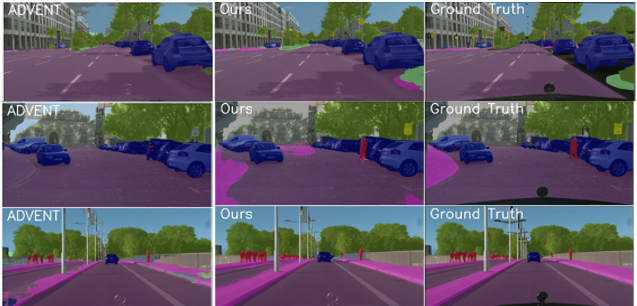}
    \caption{\textbf{Qualitative Results.} %
    We compare our results with the baseline method without baseline method (AdvEnt) \cite{vu2019advent}.}
    \label{fig:qualtitave_rec_syn2city}
\end{figure}

Figure \ref{fig:qualtitave_rec_syn2city} illustrates the qualitative results our experiments. Our new method achieves better results than without adaptation.
The continuous learning structure allows our model to continue the learning process, so improvements are made across all classes. 
The predictions produced by the model more closely matches the ground truth than the baseline without adaptation.
The noise is reduced, the edges are clearly defined, and the structure of predictions reflects the objects in the image to a greater degree. 
Overall, the model maintains the ability to generate accurate predictions and improves these predictions during test time.

\section{Conclusions}
In this paper, we have presented a novel solution to the problem of continual domain adaptation in semantic scene segmentation. 
By using the proposed Maximum Likelihood Loss to impose the constraint of distribution shift, we have avoided the problem of catastrophic forgetting and allowed the models to continuously update and improve their performance with respect to the new target data. 
Moreover, the proposed approach does not require the presence of previous training data. Therefore, our approach ensures the privacy of the previous training data.
Experiments on the standard benchmark have shown the performance improvements of our method.

\noindent
\textbf{Acknowledgment} 
This work is supported by NSF Data Science, Data Analytics that are Robust and Trusted (DART) and Googler Initiated Research Grant. We also acknowledge the Arkansas High Performance Computing Center for providing GPUs.

{
    \small
    \bibliographystyle{ieeenat_fullname}
    \bibliography{references}
}

\end{document}